\documentclass{ieeeaccess}
\UseRawInputEncoding
\ifCLASSOPTIONcompsoc
  \usepackage[nocompress]{cite}
\else
  \usepackage{cite}
\fi
\usepackage{amsmath,amssymb,amsfonts}
\usepackage{algorithmic}
\usepackage{graphicx, subfig}
\usepackage{textcomp}
\usepackage{caption}
\usepackage{algorithm}
\usepackage{algorithmic}
\usepackage{threeparttable}
\usepackage{amsmath}
\usepackage{ragged2e}
\usepackage{adjustbox}
\usepackage{multirow}
\usepackage{array}
\usepackage{booktabs}
\usepackage{soul}
\usepackage{bm}
\usepackage{url}

\soulregister\cite7 
\soulregister\citep7 
\soulregister\citet7 
\soulregister\ref7 
\soulregister\pageref7 

\def\BibTeX{{\rm B\kern-.05em{\sc i\kern-.025em b}\kern-.08em
    T\kern-.1667em\lower.7ex\hbox{E}\kern-.125emX}}
\begin{document}
 
\history{Date of publication xxxx 00, 0000, date of current version xxxx 00, 0000.}
\doi{10.1109/ACCESS.2020.DOI}

\title{Semantic Similarity Computing Model Based on Multi Model Fine-Grained Nonlinear Fusion}
\author{\uppercase{Peiying Zhang}\authorrefmark{1}, \uppercase{Xingzhe Huang}\authorrefmark{1}, \uppercase{Yaqi Wang}\authorrefmark{1}, \uppercase{Chunxiao Jiang}\authorrefmark{2,3}, \uppercase{Shuqing He}\authorrefmark{4}, \uppercase{Haifeng Wang}\authorrefmark{4}}
\address[1]{College of Computer Science \& Technology, China University of Petroleum (East China), Qingdao 266580, China}
\address[2]{Beijing National Research Center for Information Science and Technology, Tsinghua University, Beijing 100084, China}
\address[3]{Tsinghua Space Center, Tsinghua University, Beijing 100084, China}
\address[4]{College of Computer Science \& Engineering, Linyi University, Linyi 276000, China}

\tfootnote{This work is partially supported by the National Key Research and Development Program of China under Grant 2020YFB1804800, partially supported by the Shandong Province Key Research and Development Program (Major Science and Technological Innovation Project) under Grant 2019JZZY010134, and partially supported by Shandong Provincial Natural Science Foundation under Grant ZR2020MF006.}

\markboth
{Zhang \headeretal: Semantic Similarity Computing Model Based on Multi Model Fine-Grained Nonlinear Fusion}
{Zhang \headeretal: Semantic Similarity Computing Model Based on Multi Model Fine-Grained Nonlinear Fusion}

\corresp{Corresponding authors: Shuqing He (heshuqing@lyu.edu.cn), Chunxiao Jiang (jchx@tsinghua.edu.cn) and Peiying Zhang (zhangpeiying@upc.edu.cn).}

\begin{abstract}
Natural language processing (NLP) task has achieved excellent performance in many fields, including semantic understanding, automatic summarization, image recognition and so on. However, most of the neural network models for NLP extract the text in a fine-grained way, which is not conducive to grasp the meaning of the text from a global perspective. To alleviate the problem, the combination of the traditional statistical method and deep learning model as well as a novel model based on multi model nonlinear fusion are proposed in this paper. The model uses the Jaccard coefficient based on part of speech, Term Frequency-Inverse Document Frequency (TF-IDF) and word2vec-CNN algorithm to measure the similarity of sentences respectively. According to the calculation accuracy of each model, the normalized weight coefficient is obtained and the calculation results are compared. The weighted vector is input into the fully connected neural network to give the final classification results. As a result, the statistical sentence similarity evaluation algorithm reduces the granularity of feature extraction, so it can grasp the sentence features globally. Experimental results show that the matching of sentence similarity calculation method based on multi model nonlinear fusion is 84\%, and the F1 value of the model is 75\%.

\end{abstract}

\begin{keywords}
Coarse Grained, Convolution Neural Network, Attention Mechanism, Statistical Method, Multi Model Fusion.
\end{keywords}

\titlepgskip=-15pt

\maketitle

\section{Introduction}
\label{sec:introduction}
It is undeniable that feature extraction techniques have been widely used in many fields, most of which are based on deep learning, including image processing, NLP and so on. Different from image processing, the basic semantic unit of NLP \cite{6857327, 8784245, 5613968, 4077093, 8847427, 8946281, 6646387, 7050801,7972935} is sememe, it has such characters as independent, decentralized, diversification. These features determine that the model needs to grasp the meaning of the text from the coarse-grained aspect. Researchers apply attention mechanism to NLP tasks, hoping that the model can focus on text feature extraction in a more holistic way. Zhang \textit{$ et$ $al. $} \cite{8924757} propose a novel text similarity calculation model. In this model, the semantic information and location information of words in the text are taken into account. The semantic similarity between the corresponding words is calculate respectively for the sentences after word segmentation. If the similarity of words exceeds the set threshold, the location information will be compared. Finally, the weight vector is multiplied with the word vector mapping file to get the final feature matrix of the sentence. In this feature matrix, the initial coarse-grained extraction of sentence features is realized, but the influence of high-dimensional word vector weight on sentence feature matrix is too mild. The literature of\cite{8377541}  uses self attention mechanism to encode sentences, which completes the task of extracting and enlarging the main features in the process of generating sentence feature matrix. This self attention mechanism makes the addition of weight more accurate, but it is also feature weighting based on co-occurrence words and semantics. Pinheiro \textit{$ et$ $al. $} \cite{8247077} propose a sentence similarity calculation model based on the fusion of deep learning model and statistical method. The model calculates TF-IDF vector through co-occurrence words and other information, and preprocesses sentences twice before calculation.

In this paper, a sentence similarity calculation model based on multi model nonlinear fusion is proposed. The model combines the traditional sentence similarity calculation method based on statistics, and completes the coarse-grained extraction of sentence. This calculation method realizes the overall grasp of text features from the coarse-grained aspect. The extraction of sentence feature matrix based on neural network belongs to fine-grained extraction of sentence features, while feature extraction based on statistical method belongs to coarse-grained extraction. Therefore, the proposed model can achieve the fusion of coarse-grained and
fine-grained at the same time. The main contributions and innovations of this article are summarized as follows:

1. A concept of text extraction based on granularity is proposed. The feature extraction based on statistical method is defined as coarse-grained feature extraction. We combine the coarse-grained features with the fine-grained features through the fusion of various computational methods, so as to grasp the semantics of sentences from a global perspective.

2. An improved Jaccard coefficient calculation method is proposed in this paper. The traditional Jaccard coefficient only calculates the amount of words in the intersection of sentence and word segmentation results, which ignores the influence of part of speech on sentence semantics. Nevertheless for complex parts of speech, the number of words in the intersection set cannot accurately reflect the degree of semantic similarity. Therefore, improve the Jaccard algorithm by adding weighting the part of speech.

3. Compared with the traditional deep learning model, a multi feature weighting mechanism is added to the word2vec-CNN model based on multiple features. We weight the initial feature matrix by measuring the semantic similarity between words. Compared with the direct extraction of sentence feature matrix, this weighting mechanism can highlight the key points of extraction.

\section{Related Works}
As a basic task of NLP, the performance of sentence recognition model directly determines the performance of many related tasks, such as machine translation \cite{2015Neural, Nirenburg1990A, LI2019Feature, 8399736, 7090970,6464633}, automatic text summarization \cite{Cao2019Automatic, Widyassari2020Review, Zhang2019Automatic, 7462217, 8732422, 8674894, 8925331,6489503} and text generation \cite{9091179, 8772090, 8957565, 7557594, 5469163}. Most of the traditional models are based on statistical methods. In this way, the number of co-occurrence words is used to measure the similarity of sentences. However, there exists a problem that the similarity evaluation models based on statistics only stay at the digital level and do not involve the deep level of similarity evaluation. The application of deep learning model in NLP tasks has greatly improved the accuracy of sentence recognition analysis tasks. The pre-trained high-dimensional word vector is used to express the features of the sentence, which contains not only the semantic information of the sentence, but also the context information of the sentence. However, the word vector increases the time complexity of the algorithm, which is caused by the dimension of the vector. Compared with the traditional statistical methods, word vectors generate more computing tasks. In addition, the sentence feature extraction methods based on word vector extract the sentence from a more detailed perspective, which is not conducive to grasp the features of the sentence as a whole. Researchers introduce attention mechanism into the task of sentence recognition analysis, which enables the model to focus on the extraction of sentence features from a holistic perspective.

In this paper, a sentence similarity calculation method based on multi model nonlinear fusion is proposed, combining with statistical method, deep learning model and attention mechanism. In addition, the model improves the algorithm based on the previous researches. In the following section, we introduce the sentence similarity computation based on statistical method, sentence similarity computation based on neural network and based on attention mechanism.
\subsection{Sentence similarity computation based on statistical method}
Sentence similarity calculation based on statistics is realized by counting the common features between sentence pairs. In literature \cite{Yang2012A}, Yang \textit{$ et$ $al. $} conduct a comparative study of supervised feature selection methods in statistical learning of text categorization. These methods mainly include four features: document frequency (DF), information gains (IG), mutual information (MI),and term strength (TS).Through horizontal comparison experiment, it is finally determined that IG and MI has the greatest influence and are most effective among those four features. Their methods make combination of traditional statistical methods and supervised learning, which can inspect the accuracy of text classification in case of stem segmentation.

Yang \textit{$ et$ $al. $} \cite{4724621} propose two algorithms, Sentence Element Extraction algorithm (SEE) and Candidate Feature Set Extraction algorithm (CFSE), which are proposed according to the statistical idea. The basic idea of SEE is to extract features and embed them in the grammar layer. It is based on the idea that sentence components effectively express most of the meaning of the document. SEE is provided to extract sentence elements from dependency relations. Most sentence elements of words can be gained directly from their dependency relationship in the sentence. And the algorithm will be used in CFSE to get the sentence element of each term. Thus, the extraction methods based on statistical methods have gained extensive applications, especially in the field of sentence classification. However, the sentences are generally have different length, more new words, less repetitive elements etc. Sentence classification methods based on statistical learning ideas often require a lot of energy according to the characteristics of the sentences in the specific classification task for feature extraction and selection. At the same time for the new classification scene, characteristics need to be rebuilt. The methods often have poor universality. These shortcomings limit the application of this method in sentence classification to some extent.
\subsection{Sentence similarity computation and word vector generation based on neural network.}
With the development of deep learning model, researchers use neural network model to learn sentence features and generate word vectors. Such as word2vec, ELMo (Embeddings from Language Models), and BERT. Mikolov \textit{$ et$ $al.$} \cite{articlew2v} propose a neural network with input layer, hidden layer and output layer to generate word vectors. The input of the model is one-hot encoding of words in sentences, which can be divided into CBOW and skip-gram according to different training methods. However, the word vector trained in this way lacks the context information in the sentence. Peters \textit{$ et$ $ al.$} \cite{DBLP:journals/corr/abs-1802-05365} solve the context problem of word vector, and generated word vector by BiLSTM neural network. The input of the model is obtained by char level CNN. However, this model has a large number of parameters and slow training. Another model with better performance is BERT. The model proposed by Devlin \textit{$ et$ $al.$} \cite{DBLP:journals/corr/abs-1810-04805} realizes unsupervised training, which makes it possible to train a large number of unlabeled texts. Then, based on the pre-trained high-dimensional word vector, the researchers use neural network model to extract the word vector and reduce the dimension to calculate the similarity of sentence pairs. Lang \textit{$ et$ $al.$} \cite{7814716} propose two simple and effective methods by fully combining information both applied statistics and CNN with the aim to get better performance with less time cost on classification. The two methods combine statistical features and CNN based probabilistic features to build feature vectors. They choose the Naive Bayes log-count ratios as the statistical features, and improve them to adapt to multi-class tasks. Then the vectors are used into CNN to train the logistic regression classifiers. As a result, these methods combine the advantages of neural networks with statistical methods and achieve better performance than many other complex CNN models with less time cost.

Although the CNN used for sentence extraction can extract local features in sentences with  good performance. But there still remain many issues that are overlooked. For instance, the word vector representation of each word is limited by the single word vector training method, which affects the final extracted sentence features for the next step of classification. Huang \textit{$ et$ $al. $} \cite{8982686} propose a sentence classification model of convolutional cyclic neural network based on enhanced semantic feature extraction. First, the convolution kernel with semantic features is constructed by selecting the important word sequences in different categories. This step strengthens the semantic feature extraction of the important word sequences in the sentence. Then the local features of sentences are extracted and the sequence of sentences is preserved by convolution and local pooling of word vector matrix. The local features are used as the input of the cyclic neural network to obtain the long distance dependent information. The model enhances the ability of semantic feature extraction and combines the advantages of CNN and cyclic neural network. The model is effective in different categorization tasks, such as sentence level emotion classification and problem classification.
\subsection{Sentence similarity computation based on attention mechanism}
Even though sentence similarity computation technology has developed relatively mature. Many methods ignore that under specific scenes, different words in a sentence have different effects on the result of classification. There is often correlation information between words in different parts, and that information is often inevitable for the next step in classification or recognition tasks. In this sense, feature extraction methods based on attention mechanism can solve his problem to a large extent. Li \textit{$ et$ $al. $} \cite{8217843} propose an architecture of CNN with attention mechanism for chemical-induced disease extraction. It works for extracting relationships between chemicals and diseases from unstructured literature, which is of great significant to many biomedical applications such as pharmacovigilance and drug repositioning. The novel architecture for Chemical-induced disease (CID) extraction which integrates CNN, domain knowledge, the attention mechanism and piecewise strategy together. The attention mechanism is applied in pooling layer in the CNN to measure the importance features, which has shown outstanding extraction performance. In literature\cite{9023758}, the Self-Attention mechanism participates in the feature extraction process of numerous cybersecurity data existing in textual sources. With the combination of CNN and BiLSTM-CRF model, the authors propose a self-attention based neural network model for the named entity recognition in cybersecurity. First, the CNN model extracts character feature of each word in the sentence. Then the concatenation of the word embedding and the character embedding is input to the proposed model. It is worth noticing that the self-attention mechanism is added between Bi-LSTM layer and CRF layer, which can better obtain context representation of the current word and obtain more information on the current word. The authors creatively add the self-attention mechanism to capture more related feature information for the current word from the sentence itself. And experiments have fully illustrated the performance improvement because of the self-attention mechanism.

\section{The Proposed Model}
In this section, the multi model nonlinear fusion algorithm is described. The proposed model combines the sentence similarity calculation model based on statistics and the sentence similarity calculation model based on deep learning. Among them, statistical based algorithms include IF-IDF algorithm and Jaccard similarity evaluation mechanism based on part of speech. In the deep learning model, CNN is used to extract the weighted feature matrix. In the following section, the model is introduced in detail.
\subsection{Three sentence similarity computing models}
\subsubsection{SENTENCE SIMILARITY CALCULATION BASED ON TF-IDF}
Term Frequency (TF) and Inverse Document Frequency (IDF) are indicators to measure words in text. TF represents the number of times a word appears in the text, which reflects the importance of the word to the file. However, although many common words are frequently used, they do not have specific meaning or direction, so they cannot reflect the theme of the article. So the concept of IDF is introduced. The IDF represents the frequency of words appearing in a single document. The influence of common words on text theme evaluation is filtered out by taking the document as the basic unit. The combination of TF and IDF can be used to eliminate the influence of common words on semantic evaluation and evaluate the document theme better. In this paper, the sentence is used as the basic unit to calculate the IDF of words, and calculate the TF based on the number of times the word appears in the sentence pair. Then, use TF and IDF to construct TF-IDF vector of sentence pair, and calculate the similarity between sentence pairs by calculating the cosine distance between TF-IDF vectors. The calculation of the TF-IDF vector of the sentence is shown in Formula (\ref{tf-idf}).
\begin{small}
\begin{equation}
\label{tf-idf}
TF-IDF(w_i)=\frac{trem(w_i)}{Num(Sen_A \cup Sen_B)} \times \log(\frac{|T|}{1+{w_i:w_i \subset T}}),
\end{equation}
\end{small}where, $trem(w_i)$ is the number of times the word $w_i$ appears in the sentence, $Num(Sen_A \cup Sen_B)$ is the number of words shared by $Sen_A$ and $Sen_B$, $|T|$ is the total number of sentences in the dataset, $1+{w_i:w_i \subset T}$ is the number of sentence pairs containing the word $w_i$.
\subsubsection{JACCARD SIMILARITY COEFFICIENT BASED ON SENTENCE COMPONENTS}
Jaccard similarity coefficient is used to measure the similarity and difference between two samples, and the coefficient value serves as the similarity measurement result. In the task of sentence pair similarity evaluation, Jaccard similarity coefficient gives the result of sentence similarity by measuring the number of words in the intersection and union of sentence pair segmentation results. This traditional method only considers the number of co-occurrence words in the sentence pair, ignoring the influence of part of speech on sentence semantics.

In this paper, the Jaccard coefficient is weighted based on the word components. After word segmentation analysis, a sentence tree is obtained. Different sentence components have different effects on sentence semantics. We mainly consider the subject, predicate, object, attribute, adverbial and complement to measure semantic similarity. If the result set after the participle contains the subject, predicate, object, attribute, adverbial and complement in the sentence, the Jacacrd similarity coefficient will be weighted. The specific calculation formula is shown in Formula (\ref{jaccard}).
\begin{equation}
\label{jaccard}
Jaccard\_Sim=\frac{\alpha (Sen_A \cap Sen_B)}{Sen_A \cup Sen_B},
\end{equation}where, $Sen_A$ represents the segmentation result set of $Sentence_A$, and $Sen_B$ represents the segmentation result set of $Sentence_B$. $\alpha$ is the weight coefficient given according to the part of speech of overlapping words in the intersection set, and the value of $\alpha$ is given by experiment.

In order to express it more clearly, the calculation formula of Jaccard coefficient is proposed based on part of speech. Algorithm \ref{jaccardsim} is to further describe it.
\begin{algorithm}[!h]
	\caption{$Calculation$ $of$ $Jaccard$ $Similarity$ \\ $Coefficient$ $Based$ $on$ $Part$ $of$ $Speech$}
    \label{jaccardsim}
	\begin{algorithmic}[1]
        \STATE {\bf Input: }$Sentence$ $pairs$ $to$ $be$ $evaluated$, $Sen_A$ $Sen_B$;
        \STATE {\bf Output: }$Jaccard$ $similarity$
        \STATE{$Sen_A$, $Sen_B$, $Sim_{jaccard}$, $Com_{word}$, $\alpha$}
        \STATE{The word segmentation, part of speech tagging and sentence component analysis are carried out to obtain the segmentation result set with part of speech}
        \FOR{$i \in Sen_A$ }
        \FOR{$j \in Sen_B$}
        \IF{$i==j$}
        \STATE{$Com_{word}.add(i)$}
        \ENDIF
        \ENDFOR
        \ENDFOR
        \FOR{$k \in Com_{word}$}
        \IF{$K$ $satisfies$ $the$ $grammatical$ $condition$}
        \STATE{$Use$ $Algorithm$ $\ref{weight}$ $to$ $calculate$ $the$ $\alpha$}
        \ENDIF
        \ENDFOR
        \STATE{$Sim_{jaccard} = \alpha \times Com_{word} / (Sen_A \cup Sen_B)$}
        \RETURN{$Sim_{jaccard}$}
	\end{algorithmic}
\end{algorithm}
Where, $Com_{word}$ represents the common word in $Sen_{A}$ and $Sen_{B}$. $\alpha$ represents the weighted coefficient given according to the grammatical relations in co-occurrence words.

It is worth noticing that the weight coefficient is added in Formula (\ref{jaccard}) to adjust the influence of co-occurrence words on sentence similarity from the grammatical level. This effectively solves the problem that the traditional Jaccard coefficient only considers the co-occurrence of words without
considering the effect of sentence components on semantics. After getting the co-occurrence words of the sentence pairs, the syntax based Jaccard coefficient calculation model weights the sentences. Their Jaccard similarity coefficient are weighted according to the grammatical information of the co-occurrence words and the sentence components of co-occurrence words in the corresponding sentences. As for the sentence components, we mainly consider the influence of subject, predicate, attribute, adverbial and complement on sentence meaning. However, these sentence components can be arranged in a variety of ways. Only when the number of words in the co-occurrence word set is greater than 3, will the weighting mechanism be called. In addition, the effect of co-occurrence words with different components on semantics is very slight. Therefore, the model only weights the co-occurrence words when they are the same component in the sentence. The calculation method of weight coefficient is shown in Formula (\ref{weightformulation}). In Algorithm \ref{weight}, further description of the weighting mechanism is described.
\begin{equation}
\label{weightformulation}
\alpha=\frac{count+1}{count},
\end{equation}
where $count$ represents the number of words with the same sentence component in the co-occurrence word set of the sentence pair.
\begin{algorithm}[!h]
	\caption{$Sentence$ $Component$ $Weighting$ $Algorithm$}
    \label{weight}
	\begin{algorithmic}[1]
        \STATE {\bf Input: }$Com_{word}$;
        \STATE {\bf Output:} $the$ $value$ $of$ $weight$
        \STATE{$count$, $\alpha$}
        \IF{$length(Com_{word}) > =3$}
        \FOR{$word_{i} \in Com_{word}$}
        \STATE{$word_{i}$ has the same grammatical component in $Sen_{A}$ and $Sen_{B}$}
        \STATE{$count+=1$}
        \ENDFOR
        \IF{$count==0$}
        \RETURN{$\alpha=1$}
        \ELSE
        \RETURN{$\alpha=(count+1)/count$}
        \ENDIF
        \ELSE
        \RETURN{$\alpha=1$}
        \ENDIF
	\end{algorithmic}
\end{algorithm}
\subsubsection{WORD2VEC-CNN ALGORITHM BASED ON MULTIPLE FEATURES}
Traditional models overlook the mutual information between text pairs before they are input into the deep learning model, which makes it impossible for the model to effectively extract the association information in the text when extracting features. Instead, word2vec-CNN algorithm we present takes multiple features into account, which is a sentence feature extraction model based on attention mechanism. The model constructs a multi feature attention matrix based on the co-occurrence word information, location information and semantic information between texts to realize the weighting of the original feature matrix. After the word segmentation mapping of the sentence, the feature vectors representing $Sen_A =  w_{1}, w_{2},...,w_{n} $ and $Sen_B = {w_{1}, w_{2},...,w_{m}}$ are obtained.
\begin{equation}
\label{cos}
Sim_{word}= \sum_{i}^n\sum_{j}^mCOS(w_{i},w_{j}),w_i \in Sen_A, w_j \in Sen_B,
\end{equation}
where, $n$ and $m$ represent the length of $Sen_A$ and $Sen_B$ respectively. The cosine distance of word vector between sentence pairs is calculated, and the calculation method is shown in Formula \ref{cos}. The word vector matrix of sentence pairs is calculated. The calculation method as shown in Formula (\ref{matrixembedding}).
\begin{equation}
\begin{split}
\label{matrixembedding}
w2v\_matrix=\sum_{i}^{n} \sum_{j}^{m}COS(w_i,w_{j}^{,})\\ =\left[                 
  \begin{array}{cccc}   
    COS(w_1,w_{1}^{,}),&...&,COS(w_1,w_{j}^{,})\\  
    COS(w_2,w_{1}^{,}),&...&,COS(w_2,w_{j}^{,})\\
    COS(w_3,w_{1}^{,}),&...&,COS(w_3,w_{j}^{,})\\
    ...,&...&,... \\
    COS(w_i,w_{1}^{,}),&...&,COS(w_i,w_{j}^{,})\\
  \end{array}
\right]  ,
\end{split}
\end{equation}
After getting the weight matrix between sentence pairs, the weight vector of each unit in $Sen_A$ relative to $Sen_B$ is calculated by summing the row elements of the matrix. The matrix column elements are summed to calculate the weight vector of each semantic unit in $Sen_B$ relative to $Sen_A$. Take the word position into account and generate the position embedded weight matrix according to the editing distance of the words in the sentence. Firstly, the co-occurrence words in the sentence are retrieved globally and a set of co-occurrence words is generated, where $k$ is the number of co-occurrence words in the text, $w_k^c \in set(A) \cap set(B)$. Then, the position information of co-occurrence word $w_k^c$ in $Sen_A$ is retrieved from the sentence pair, which is marked as $loc_A(w_k^c)$. In the same way, the position information in $Sen_B$ is recorded as $loc_B(w_k^c)$. Finally, the position information is used as an index to obtain the words of corresponding positions in the sentence pair, and the edit distance between them and co-occurrence words is calculated to generate the position embedding weight matrix based on the editing distance. The calculation process is shown in Formula (\ref{posembedding}).
\begin{equation}
\label{posembedding}
pos\_embedding=\left\{
\begin{aligned}
\frac{2*Editdistance(w_k^{c},w_{loc_A({w_k^{c}})}^{,})}{\min{\{length(A),length(B)}\}}, \\ loc_A(w_k^c) \le length(B)\\
0, other \\
\frac{2*Editdistance(w_k^{c},w_{loc_B({w_k^{c}})}^{,})}{\min{\{length(A),length(B)}\}}, \\ loc_B(w_k^c) \le length(A)
\end{aligned}
\right.
\end{equation}
The number of words in the sentence and the corresponding position also affect the semantic changes. Position embedding generates a position embedding weight matrix based on the edit distance of words in the text. First, the co-occurrence words in the text are retrieved globally and a co-occurrence word set is generated $set_{comWord}=\{w_1^{c},w_2^{c},...,w_k^{c}\}$, where $k$ represents the number of co-occurrence words in the sentence, $w_i^{c}\in set(A)\cap set(B)$. Then, the position information of co-occurrence words $w_k^{c}$in $sentence_A$ is retrieved in the text pair, which is recorded as $loc_A(w_k^{c})$, and the position information in $sentence_B$ is recorded as $loc_B(w_k^{c})$. Finally, the position information is used as the index to obtain the words corresponding to the position of the text pair, and the edit distance between the words and the co-occurrence words is calculated to generate the position embedding weight matrix based on the edit distance. The calculation process is shown in Formula (\ref{posembedding}).

After calculating the weight vector based on word2vec embedding and the position embedding vector based on edit distance, the row vector and column vector are added respectively. Then, the probability normalization is performed by softmax function. The generated normalized vector is used to weight the text matrix to get the input of neural network. See Formula (\ref{attationmatrix1}) and Formula (\ref{attationmatrix2}) for calculation.
\begin{equation}
\begin{split}
\label{attationmatrix1}
Att\_Matrix_{Sen_A}= softmax(row\_vec+pos\_row)* \\ [w_1,w_2,...,w_n],
\end{split}
\end{equation}

\begin{equation}
\begin{split}
\label{attationmatrix2}
Att\_Matrix_{Sen_B}= softmax(col\_vec+pos\_col)* \\ [w_1^{,},w_2^{,},...,w_n^{,}]^T,
\end{split}
\end{equation}

among them, $Att\_Matrix_{Sen_A}$ and $Att\_Matrix_{Sen_B}$ represent the multi feature attention sentence matrix after weighting the mutual information in the sentence. After obtaining the weighted matrix representation of the sentence pair, the next step is to input the sentence pairs into the convolution neural network for sentence feature extraction and obtain their feature vector representation. The convolution neural network only includes convolution layer and pooling layer. The overall structure of the model is shown in FIGURE \ref{word2vec-cnn}.

\begin{figure}[ht]
\centering
\includegraphics[scale=0.47]{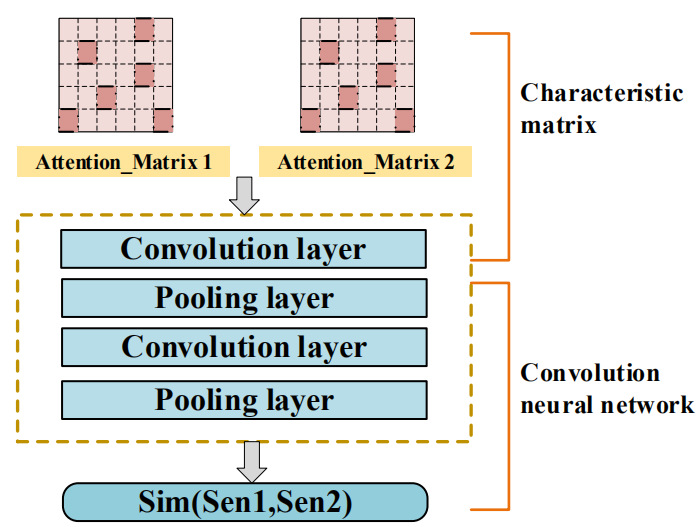}
\caption{The structure of word2vec-cnn.}
\label{word2vec-cnn}
\end{figure}
\subsection{Sentence similarity computing model based on multi model nonlinear fusion}
In order to give full play to the advantages of the above three models, a sentence similarity calculation algorithm based on multi model nonlinear fusion is proposed. In this algorithm, TF-IDF algorithm, Jaccard coefficient algorithm based on part of speech and word2vec-CNN algorithm based on multi feature are used to calculate the sentence similarity scores. After normalization and weighting, the sentence similarity scores are input into a shallow neural network to learn the feature relations in the output results. Then the final calculation results of sentence similarity are given. The weights used in the calculation are obtained from the performance evaluation parameters of the three models, as shown in Formula (\ref{fuse}).

\begin{equation}
\label{fuse}
[\alpha, \beta, \gamma] = sigmoid(Jaccard, word2vec-CNN, IF-IDF).
\end{equation}
The overall structure of sentence similarity calculation algorithm based on multi model nonlinear fusion is shown in FIGURE \ref{word2vec-cnn}.
\begin{figure}[ht]
\centering
\includegraphics[scale=0.48]{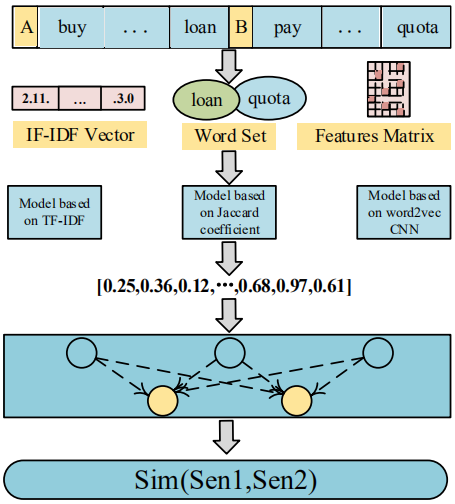}
\caption{The structure of model.}
\label{word2vec-cnn}
\end{figure}

\section{Experiment and Result Analysis}
In this section, in order to verify the performance of the model, we use ant financial NLP competition data set (\url{https://dc.cloud.alipay.com/index?click_from=MAIL&bdType=acafbbbiahdahhadhiih#/topic/intro?id=3}) and semantic\_textual\_similarity (\url|http://nlpprogress.com/english/semantic_textual_similarity.html|) data set to verify the model respectively.
\subsubsection{ant financial data set}
The ant financial data set comes from the Chinese text similarity contest held by Alipay. Its data mainly come from Alipay's customer service data. Each pair of sentences is marked by a unique similarity classification result, where 0 represents the same semantics and 1 represents different semantics. In addition, the training set contains 100000 pieces of data, and the test set contains 10000 pieces of data. The standard format of the dataset is shown in TABLE \ref{dataset}, where, $1^*$, $2^*$, $3^*$ represent the translation of Chinese sentences. It is also in TABLE \ref{concretesentencepairs}. The ratio of positive and negative samples is 1:4.
\subsubsection{semantic\_textual\_similarity data set}
In order to compare with the related models, semantic\_textual\_similarity (STS) datasets from 2012 to 2015 in the experiment are also used. STS dataset divides data from different fields into different documents, including the semantic similarity scores of sentence pairs, which range from 0 to 5. Compared with the data of ant financial, STS dataset divides the semantic similarity of sentence pairs more carefully. It is worth noting that we multiply the similarity between 0 and 1 of the model output by 5 and compare it with the labels in the dataset.
\begin{table*}[!htbp]
\caption{ Dataset format of the input model.}
\label{dataset}
\centering
\scalebox{1.4}{
      \begin{tabular}{ccccccc}
      \hline
       Num & Sen1 & Sen2 & Sim \\ \hline
       $ $ $ $ $1^{*}$ & Ant credit pay is locked after the deadline. & Ant credit pay has no repayment and is locked & 0\\
       $ $ $ $ $2^{*}$ & Ant credit pay is limited. & Ant credit pay can't be used. & 0 \\
       $ $ $ $ $3^{*}$ & The quota of ant credit pay is locked & The loan limit of ant credit pay is not enough & 0 \\ \hline
      \end{tabular}
      }
\end{table*}
The experimental platform of this paper is windows 10 operating system, using pycharm as the development tool. Based on tensorflow 2.0, a deep learning model is built. The model effect analysis is carried out by comparing with the experimental results of similar works. During training, the triples containing sentence pairs and tags are input into the model. The input data format is $(sentence1, sentence 2, 0 / 1)$. Suppose that $score$ represents the sentence pair score. If $| score-0 | > | score-1 |$, the sentence pair is considered to be semantically different. If $| score-0 | < | score-1 |$, the semantic similarity is considered. In the experiment, $accuracy$, $precision$, $recall$ and $F_1$ are used as the evaluation criteria of the model. The specific calculation formula is as follows:

\begin{equation}
accuracy = \frac {TP+TN}{TP+TN+FP+FN},
\end{equation}

\begin{equation}
precision = \frac {TP}{TP+FP},
\end{equation}

\begin{equation}
recall = \frac {TP}{TP+FN},
\end{equation}

\begin{equation}
F_1 = \frac {2 \times Precision \times Recall}{Precision+Recall},
\end{equation}
where TP means that the actual result is positive and the predicted result is positive, TN means that the actual result is negative and the predicted result is negative, FP means that the actual result is negative while the predicted result is positive, FN means that the actual result is positive and the predicted result is negative.
\subsection{Performance comparison of multiple models}
The method based on multi model nonlinear fusion considers TF-IDF algorithm, part of speech based Jaccard coefficient algorithm and word2vec-CNN algorithm. The three algorithms calculate the similarity of problem pairs from the perspective of word frequency and context relevance. To compare the performance of the three models in calculating similarity, the accuracy, recall rate and other evaluation parameters were selected for an intuitive comparison. The three algorithms are compared on the ant financial training set. As can be seen from TABLE \ref{multimodelsresult}, word2vec-CNN model has the highest accuracy. This is due to the rich feature information contained in the training high-dimensional word vector. Compared with the deep learning model, the algorithm based on Jaccard coefficient can save computing resources and achieve an accuracy of 0.79. As illustrated in the TABLE \ref{multimodelsresult}, the accuracy of TF-IDF is the worst, and it is not significant to calculate the word frequency in a short sentence pair sequence. A single count of the occurrence times of words is more suitable for converting the weight to the weight of the original sentence. Based on the experimental results in TABLE \ref{multimodelsresult}, the output of each model is weighted and normalized before input to the fully connected neural network of multi model fusion. (\textbf{Note that the function of TABLE \ref{multimodelsresult} is to determine the weighting coefficient according to the experimental data of the three models, not the final output of the models.})
\begin{table}[!htbp]
\caption{ Comparison of experimental results of three models.}
\label{multimodelsresult}
\centering
\scalebox{1.14}{
      \begin{tabular}{ccccccc}
      \hline
       models & accuracy & precision & recall & $F_1$ \\ \hline
       TF-IDF & 0.25 & 0.22 & 0.97 & 0.36 \\
       Jaccard & 0.79 & 0.54 & 0.11 & 0.18 \\
       word2vec-CNN & 0.80 & 0.53 & 0.82 & 0.65 \\ \hline
      \end{tabular}
      }
\end{table}
\subsection{Performance comparison of models with different weights}
In order to verify the performance index of the model under different weights, the weighted vectors are constructed by accuracy, precision, recall and $F_1$ values, and then multiplied with the output of each model. After weighting precision, accuracy and $F_1$ value of the multi model nonlinear fusion algorithm are improved significantly. From the experimental results in TABLE \ref{multimodelsresult}, it can be noted that the accuracy of the model based on TF-IDF is much lower than that of the other two models, so an appropriate weighting factor should be chosen to reduce its influence. After normalization of the experimental results of the three models, the reliability of the calculated results is shown in TABLE \ref{weightresult}. Selecting the accuracy to generate weighting factor can keep the calculation results of the part of speech based Jaccard algorithm and word2vec-CNN algorithm to the maximum extent, and reduce the influence of TF-IDF algorithm on multi model nonlinear fusion algorithm. The way to generate weighting factors based on precision and recall has little effect on the accuracy and $F_1$ value of the model. However, the calculation results are still better than the unweighted method. $F_1$ value is an overall evaluation parameter of model performance, which is slightly less accurate compared with the way of generating weighting factors with accuracy. But it improves the overall performance of multi model nonlinear fusion algorithm. The algorithm based on multi model nonlinear fusion can select the way to generate weighting factors according to the needs of different tasks, so as to improve the adaptability of the model as low as possible according to the characteristics of the data set. In the next experiment, set the $\alpha=0.38$, $\beta=0.40$, $\gamma=0.22$.
\begin{table}[!htbp]
\caption{ Comparison of experimental results of multi feature nonlinear fusion model with different weighting factor. When the weighting factor is accuracy, it means that the accuracy of the three models is used to normalize and calculate the weight vector. The normalization method is shown in Formula \ref{fuse}.}
\label{weightresult}
\centering
\scalebox{1.12}{
      \begin{tabular}{ccccccc}
      \hline
       weighting factor & $\alpha$ & $\beta$ & $\gamma$ & $accuracy$ & $F_1$ \\ \hline
        -- & -- & -- & -- & 0.77 & 0.65 \\
       accuracy & 0.38 & 0.40 & 0.22 & 0.84 & 0.75 \\
       precision & 0.37 & 0.36 & 0.27 & 0.78 & 0.65 \\
       recall & 0.19 & 0.38 & 0.43 & 0.79 & 0.70 \\
       $F_1$ & 0.26 & 0.42 & 0.32 & 0.82 & 0.79 \\ \hline
      \end{tabular}
      }
\end{table}
\subsection{Comparison of related models}
In order to explain the performance of the model, In Experiment 1, we used ant financial data set, and compared the proposed model with three related works, which mainly used TF-IDF, CNN and so on. The purpose is to verify the advantages and disadvantages of MMNF by comparing with similar models. In Experiment 2, STS dataset was used. The comparison with some state-of-art models like BERT are finished. The purpose is to prove the significance of the work by comparing with the mainstream models.
\subsubsection{Experiment 1}
In Experiment 1, comparisons with the traditional models are completed.

$\bullet$ \textbf{TF-IDF}: VBS \cite{10.5555/1597538.1597662} (vector based similarity) is a calculation method of vector representation similarity based on TF-IDF weight. It was proposed by Mihalcea in 2006. This method uses TF-IDF to obtain sentence vector, which is a statistical method to obtain word vector.

$\bullet$ \textbf{Weighted w2v}: Gao \textit{$ et$ $ al.$} \cite{0Text} proposed a short text classification method based on word2vec weighting, which adopted the same training method of word vector as the model in this paper.

$\bullet$ \textbf{w2v-CNN}: In literature \cite{GaoText}, a CNN model is proposed to process word vector for text similarity clustering. The model combines neural network model and document topic model Latent Dirichlet allocation (LDA) to construct feature matrix. The matrix model can not only effectively represent the semantic features of the words but also convey the context features and enhance the feature expression ability of the model.

\begin{figure}[h]
\centering
\includegraphics[height=6cm,width=8cm]{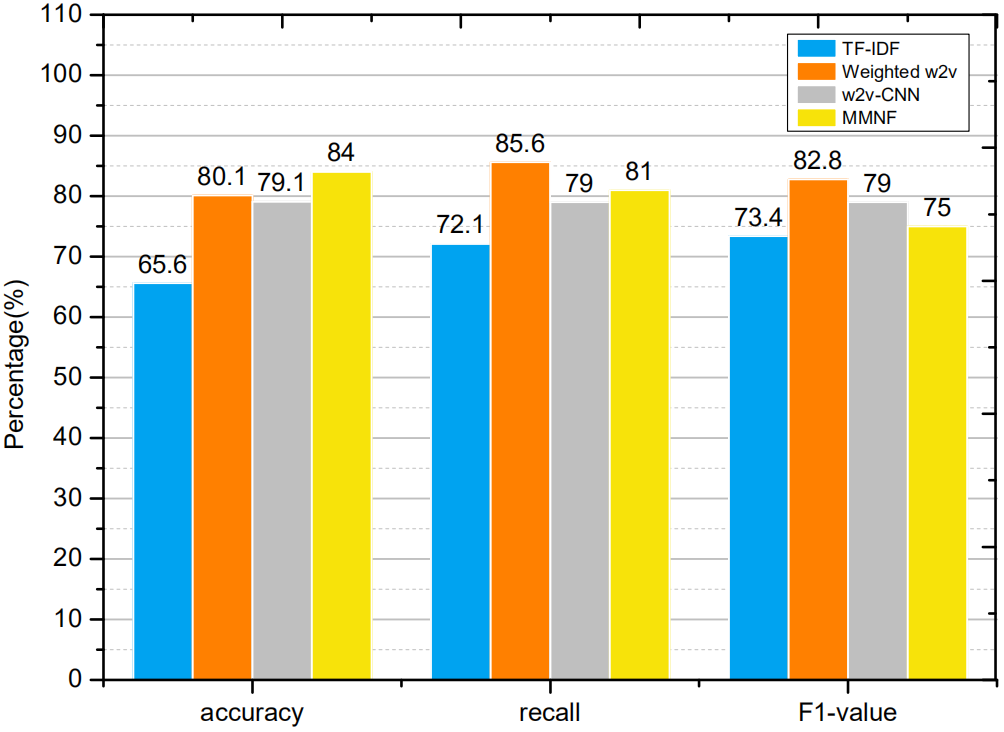}
\caption{Correlation model comparison.}
\label{comparison}
\end{figure}

In FIGURE \ref{comparison}, the accuracy of the model is compared and the experimental results are analyzed. From the data in the diagram, it can be concluded that the model based on TF-IDF has poor effect, mainly due to the fact that the model considers too single factors, only considering the frequency of words in sentences and only having one single measurement factor correspondingly. Compared with TF-IDF model, the model based on word vector has demonstrates its superiority in accuracy, recall rate and F1-value.

\begin{table*}[!htbp]
\caption{ Similarity calculation of concrete sentence pairs.}
\label{concretesentencepairs}
\centering
\scalebox{0.85}{
      \begin{tabular}{ccccccc}
      \hline
       index & $Sen_1$ & $Sen_2$ & $Jaccard$ & $word2vec-CNN$ & $TF-IDF$ & $MMFC$ \\ \hline
       $ $ $ $ $1^{*}$ & Can I use ant credit pay. & Can't I use ant credit pay. & -- & -- & -- & -- \\
       $ $ $ $ $2^{*}$ & What's the minimum amount of ant credit pay. & What is the available quota of my ant credit pay. & -- & -- & -- & -- \\
       $ $ $ $ $3^{*}$ & The ant credit pay are emptied by stages. & Ant credit pay query by stages. & -- & -- & -- & -- \\
       $ $ $ $ $4^{*}$ & The quota of ant credit pay is locked. & The loan limit of ant credit pay is enough. & -- & -- & -- & -- \\ \hline
      \end{tabular}
      }
\end{table*}

\subsubsection{Experiment 2}
In Experiment 2, we mainly compare the model based on BERT.

$\bullet$ \textbf{BERT[CLS]} (Devlin \textit{$ et$ $al. $} \cite{DBLP:journals/corr/abs-1810-04805}): In the text classification task, the BERT model adds a [CLS] token to each input sentences, where [CLS] represents the meaning of classification. In the process of sentence calculation in the BERT model, since [CLS] does not have any meaning of words, it will integrate the meanings of each word more evenly. Finally, the [CLS] vector is the feature vector of the sentence.

$\bullet$ \textbf{Average BERT} (Devlin \textit{$ et$ $al. $} \cite{DBLP:journals/corr/abs-1810-04805}): It is different from [CLS], average BERT model is to average the representations obtained from the last layer.

$\bullet$ \textbf{Sentence-BERT} (Reimers \textit{$ et$ $al. $} \cite{DBLP:journals/corr/abs-1908-10084}): BERT model has set a new state-of-the-art performance in many NLP tasks. However, it causes a massive time consumption. Sentence-BERT (SBERT) reduces the time consumption of semantic similarity calculation task. This model adds siamese and triplet network structures to derive semantically meaningful sentence embedding on the basis of the BERT model.
\begin{table}[tp]
\centering
\scalebox{1.0}{
\begin{threeparttable}
\caption{Experimental result on STS dataset in terms of the Pearson correlation coefficients (left, *100)  and Spearman rank correlation coefficients (right, *100), where the best result are shown in bold face.}
\label{bertcomparison}
\begin{tabular}{c|ccc|c}
\toprule
\multirow{1}{*}{Year}&
\multicolumn{3}{c}{Compared methods}&Our method\cr
 & BERT [CLS] &Average BERT & SBERT & MMNF \cr
\midrule
2012& 27.5 / 32.5 & 46.9 / 50.1 & 64.6 / 63.8& \bm{${65.2}$} / \bm{${64.3}$} \cr
\hline
2013&22.5 / 24.0&52.8 / 52.9&67.5 / \bm{${69.3}$}& \bm{${68.3}$} / 67.9 \cr
\hline
2014&25.6 / 28.5&57.2 / 54.9& \bm{${73.2}$} / \bm{${72.9}$} & 69.4 / 70.2\cr
\hline
2015&32.1 / 35.5&63.5 / 63.4& \bm{${74.3}$} / \bm{${75.2}$}& 69.1 / 68.7 \cr
\bottomrule
\end{tabular}
\end{threeparttable}}
\end{table}

In the experiment, comparison are made with three models based on BERT. The experimental results are shown in TABLE \ref{bertcomparison}. From the table, we can see that the BERT [CLS] model leads to rather poor performances. The Bert model is a language model based on pre-training and fine-tuning. The output [CLS] vector is directly used as the output of the whole sentence in the BERT [CLS] model, which makes it difficult for the model to be suitable for text classification tasks. The experimental data of average BERT model is much better than that of BERT [CLS], which indicates that CLS token can not extract all the words in sentences well. In the proposed model, the pre-trained vector is used, and the multi feature mechanism is used for fine-tuning. The structure of the model is similar to that of the BERT model. The difference is that this model further enhance the output characteristics of word2vec model, and improve the effect of the model by multi model fusion. The experimental results are far better than that of BERT[CLS] and average BERT. SBERT model has shown satisfactory results in 2013-2015 experimental data, which once again demonstrates the advantages of the BERT model and its pre-training and fine-tuning structure. But in the data of 2012, the data of our model is better than that of SBERT. Only relying on the Bert model to extract features may lead to the loss of features to a certain extent. The superposition of features based on the Bert model may further improve the performance of the model. And further strengthening the features of the pre-training model may further improve the accuracy of NLP tasks, which will be future research work.

\subsection{Similarity calculation of concrete sentence pairs}
In the selected data set, sentences with various structures are included. The computational characteristics of different models may be suitable for sentence pairs with different structures. For example, the improved Jaccard algorithm is suitable for sentence pairs with more co-occurrence words. However, there remain also cases in which the semantic meaning of sentences containing many co-occurrence words is exactly opposite. The multi-model nonlinear fusion algorithm understands the relationship between data from various calculation results, which enables the model to weigh different calculation results, so as to achieve the purpose of generating classification results more in line with the actual sentence meaning. In TABLE \ref{concretesentencepairs}, the results of processing special sentence pairs by different models and the calculation results of multi model nonlinear fusion algorithm are illustrated. Four sentences are chosen for comparison, with the sentence pair 2, sentence pair 3 and sentence pair 4 containing more co-occurrence words. The improved Jaccard algorithm gives higher similarity based on the principle of counting co-occurrence words. But the semantics of these two pairs of sentences are different. In contrast, the similarity scores calculated by word2vec-CNN and TF-IDF are lower. The multi model nonlinear fusion algorithm realizes the correct classification of sentence semantics based on the comprehensive evaluation of three models. Sentence pair 1 has the same meaning, but the number of co-occurrence words in the text is not large. Therefore, Jaccard algorithm and TF-IDF algorithm both get low similarity score. But the multi model nonlinear fusion algorithm based on word2vec-CNN mechanism finally gives the correct evaluation score. It can be seen that the multi model nonlinear fusion algorithm proposed in this paper can realize the trade-off between the calculation results of various models, and the calculation results are the most accurate.

\section{Conclusion and Future Work}
In this paper, a multi model nonlinear fusion algorithm is proposed for different sentence structure features. The results of Jaccard algorithm, TF-IDF algorithm and word2vec-CNN are input into the shallow fully connected neural network to train the model and give an ideal classification result. The improved Jaccard algorithm takes the grammatical information into account in the calculation process of similarity, so that the single feature based on the number of co-occurrence words is supplemented. TF-IDF algorithm calculates sentence similarity from word frequency and inverse document frequency. In word2vec-CNN algorithm, sentence feature matrix is weighted by multi feature attention mechanism. The algorithm of multi model nonlinear fusion can comprehensively judge from the calculation results of each model, synthesize the superiority of three kinds of models and give a reasonable calculation result, which improves the calculation accuracy.

The experimental results show that the proposed sentence similarity calculation method based on multi feature fusion can balance the calculation results of multiple models. However, there is still room for improvement. The word vector based on word2vec can not express the context of the text. In different contexts, the same word may have different meanings. The word vector given by word2vec model is static and cannot describe the dynamic change of semantics. In the future, we will use other language models to improve our work, such as ELMo, BERT, etc.
\section*{Acknowledgments}
This work is partially supported by the National Key Research and Development Program of China under Grant 2020YFB1804800, partially supported by the Shandong Province Key Research and Development Program (Major Science and Technological Innovation Project) under Grant 2019JZZY010134, and partially supported by Shandong Provincial Natural Science Foundation under Grant ZR2020MF006.

\ifCLASSOPTIONcaptionsoff
  \newpage
\fi



%
%
%

\bibliography{references}

%

\begin{IEEEbiography}[{\includegraphics[width=1in,height=1.25in,clip,keepaspectratio]{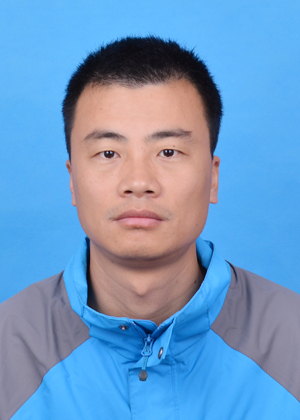}}]{Peiying Zhang}
is currently an Associate Professor with the College of Computer Science and Technology, China University of Petroleum (East China). He received his Ph.D. in the School of Information and Communication Engineering at University of Beijing University of Posts and Telecommunications in 2019. He has published multiple IEEE Trans./Journal/Magazine papers since 2016, such as IEEE TNSE, IEEE TNSM, IEEE TVT, IEEE TETC, IEEE IoT-J, COMPUT COMMUN, IEEE COMMUN MAG, and etc. His research interests include semantic computing, future internet architecture, network virtualization, and artificial intelligence for networking.
\end{IEEEbiography}

\begin{IEEEbiography}[{\includegraphics[width=1in,height=1.25in,clip,keepaspectratio]{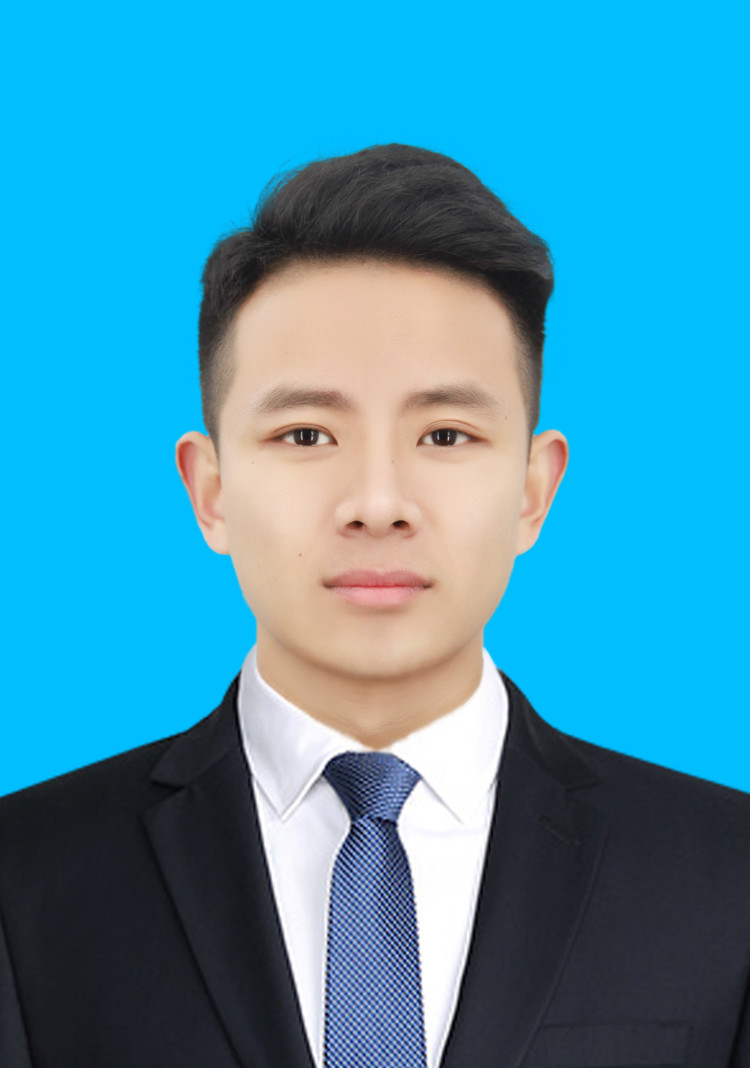}}]{Xingzhe Huang}
is currently a graduate student in the College of Computer Science and Technology, China University of Petroleum (East China). His research interests include artificial intelligence and natural language processing.
\end{IEEEbiography}
\begin{IEEEbiography}[{\includegraphics[width=1in,height=1.25in,clip,keepaspectratio]{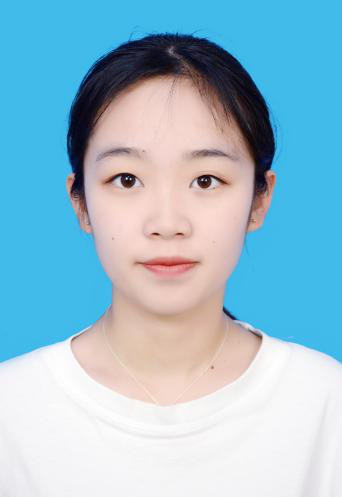}}]{Yaqi Wang}
is currently a graduate student in the College of Computer Science and Technology, China University of Petroleum (East China). Her research interests include machine learning and natural language processing.
\end{IEEEbiography}

\begin{IEEEbiography}[{\includegraphics[width=1in,height=1.25in,clip,keepaspectratio]{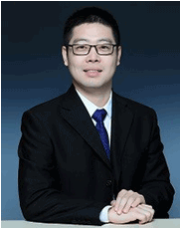}}]{Chunxiao Jiang}
is an Associate Professor in School of Information Science and Technology, Tsinghua University. He received the B.S. degree in information engineering from Beihang University, Beijing in 2008 and the Ph.D. degree in electronic engineering from Tsinghua University, Beijing in 2013, both with the highest honors. His research interests include application of game theory, optimization, and statistical theories to communication, networking, and resource allocation problems, in particular space networks and heterogeneous networks. Dr. Jiang has served as an Editor of IEEE Internet of Things Journal, IEEE Network, IEEE Communications Letters, and a Guest Editor of IEEE Communications Magazine, IEEE Transactions on Network Science and Engineering, and IEEE Transactions on Cognitive Communications and Networking. He has also served as a member of the technical program committee as well as the Symposium Chair for a number of international conferences, including IEEE ICC 2018 Symposium Co-Chair, IWCMC 2018/2019 Symposium Chair, WiMob 2018 Publicity Chair, ICCC 2018 Workshop Co-Chair, and ICC 2017 Workshop Co-Chair. Dr. Jiang is the recipient of the Best Paper Award from IEEE GLOBECOM in 2013, the Best Student Paper Award from IEEE GlobalSIP in 2015, IEEE Communications Society Young Author Best Paper Award in 2017, the Best Paper Award IWCMC in 2017, IEEE ComSoc TC Best Journal Paper Award of the IEEE ComSoc TC on Green Communications \& Computing 2018, IEEE ComSoc TC Best Journal Paper Award of the IEEE ComSoc TC on Communications Systems Integration and Modeling 2018, the Best Paper Award ICC 2019.
\end{IEEEbiography}

\begin{IEEEbiography}[{\includegraphics[width=1in,height=1.25in,clip,keepaspectratio]{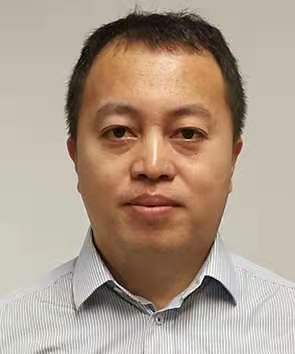}}]{Shuqing He}
received the M.S. degree in computer application technology in 2006 from China University of Petroleum (East China) in China. He is currently a Ph.D. candidate in the State Key Laboratory of Networking and Switching Technology at Beijing University of Posts and Telecommunications. He is a teacher in the School of Information Science and Technology of Linyi University. His main research interests include Service-Oriented Computing, Edge Computing and Data Analysis and Processing.
\end{IEEEbiography}

\begin{IEEEbiography}[{\includegraphics[width=1in,height=1.25in,clip,keepaspectratio]{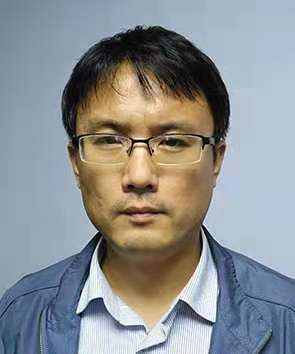}}]{Haifeng Wang}
is a professor and Director of the Data-Intensive Computing Laboratory in the Computer Science Department of Linyi University. He received the BS degree in computer application from Shandong University, Jinan, China, in 1997, the MS degree in computer application technology from China University of Petroleum, Qingdao, China, in 2005, and received the PhD degree in Computer Science from the University of Shanghai for Science and Technology, Shanghai, China, in 2012. His research interests mainly include data-intensive computing, parallel and distributed computing, and big data technology.
\end{IEEEbiography}

\EOD

\end{document}